# Active Imitation Learning via Reduction to I.I.D. Active Learning

**Kshitij Judah**
School of E.E.C.S.
Oregon State University
Corvallis, OR 97331-5501, USA
judahk@eecs.oregonstate.edu

**Alan P. Fern**
School of E.E.C.S.
Oregon State University
Corvallis, OR 97331-5501, USA
afern@eecs.oregonstate.edu

**Thomas G. Dietterich**
School of E.E.C.S.
Oregon State University
Corvallis, OR 97331-5501, USA
tgd@cs.orst.edu

## Abstract

In standard passive imitation learning, the goal is to learn a target policy by passively observing full execution trajectories of it. Unfortunately, generating such trajectories can require substantial expert effort and be impractical in some cases. In this paper, we consider active imitation learning with the goal of reducing this effort by querying the expert about the desired action at individual states, which are selected based on answers to past queries and the learner's interactions with an environment simulator. We introduce a new approach based on reducing active imitation learning to i.i.d. active learning, which can leverage progress in the i.i.d. setting. Our first contribution, is to analyze reductions for both non-stationary and stationary policies, showing that the label complexity (number of queries) of active imitation learning can be substantially less than passive learning. Our second contribution, is to introduce a practical algorithm inspired by the reductions, which is shown to be highly effective in four test domains compared to a number of alternatives.

## 1 Introduction

Traditionally, passive imitation learning involves learning a policy that performs nearly as well as an expert's policy based on a set of trajectories of that policy. However, generating such trajectories is often tedious or even impractical for an expert (e.g. real-time low-level control of multiple game agents). In order to address this issue, we consider active imitation learning where full trajectories are not required, but rather the learner asks queries about specific states, which the expert labels with the correct actions. The goal is to learn a policy that is nearly as good as the expert's policy using as few queries as possible.

The active learning problem for i.i.d. supervised learning has received considerable attention both in theory and practice (Settles, 2009), which motivates attempting to leverage that work for active imitation learning. However, the direct application of i.i.d. approaches to active imitation learning can be problematic. This is because i.i.d. active learning algorithms assume access to either a target distribution over unlabeled input data (in our case states) or a large sample drawn from it. The goal then is to select the most informative query to ask, usually based on some combination of label (in our case actions) uncertainty and unlabeled data density. Unfortunately, in active imitation learning, the learner does not have direct access to the target state distribution, which is the state distribution induced by the unknown expert policy.

In principle, one could approach active imitation learning by assuming a uniform or an arbitrary distribution over the state space and then apply an existing i.i.d. active learner. However, such an approach can perform very poorly. This is because if the assumed distribution is considerably different from that of the expert, then the learner is prone to ask queries in states rarely or even never visited by the expert. For example, consider a bicycle balancing problem. Clearly, asking queries in states where the bicycle has entered an unavoidable fall is not very useful because no action can prevent a crash. However, i.i.d. active learning technique will tend to query in such uninformative states, leading to poor performance, as shown in our experiments. Furthermore, in the case of a human expert, a large number of such queries poses serious usability issues, since labeling such states is clearly a wasted effort from the expert's perspective.

In this paper, we consider the problem of reducing active imitation learning to i.i.d. active learning both in theory and practice. Our first contribution is to analyze the PAC label complexity (number of expert queries) of a reduction for learning non-stationary policies, which requires only minor modification to exist-ing results for passive learning. Our second contribution is to introduce a reduction for learning stationary policies resulting in a new algorithm *Reduction-based Active Imitation Learning (RAIL)* and an analysis of the label complexity. The resulting complexities for active imitation learning are expressed in terms of the label complexity for the i.i.d. case and show that there can be significant query savings compared to existing results for passive imitation learning. Our third contribution is to describe a new practical algorithm, RAIL-DW, inspired by the RAIL algorithm, which makes a series of calls to an i.i.d. active learning algorithm. We evaluate RAIL-DW in four test domains and show that it is highly effective when used with an i.i.d. algorithm that takes the unlabeled data density into account.

## 2 Related Work

Active learning has been studied extensively in the i.i.d. supervised learning setting (Settles, 2009) but to a much lesser degree for sequential decision making, which is the focus of active imitation learning. Several studies have considered active learning for *reinforcement learning (RL)* (Clouse, 1996; Mihalkova and Mooney, 2006; Gil et al., 2009; Doshi et al., 2008), where learning is based on both autonomous exploration and queries to an expert. Rather, in our imitation learning framework, we do not assume a reward signal and learn only from expert queries. Other prior work (Shon et al., 2007) studies active imitation learning in a multiagent setting where the expert is itself a reward seeking agent and hence is not necessarily a helpful expert. Here we consider only helpful experts.

One approach to imitation learning is *inverse RL (IRL)* (Ng and Russell, 2000), where a reward function is learned based on a set of target policy trajectories. The learned reward function and transition dynamics are then given to a planner to obtain a policy. There has been limited work on active IRL. This includes a Bayesian approach (Lopes et al., 2009), where a posterior over reward functions is used to select a "most informative" query. Another Bayesian approach (Cohn et al., 2010) models uncertainty about the entire MDP model and uses Expected Myopic Gain (EMG) to select a query about transition dynamics and rewards. While promising, the scalability of these approaches is hindered by the assumptions made by IRL and have only been demonstrated on small problems. In particular, they require that the exact domain dynamics are provided or can be learned and that an efficient planner is available.

To facilitate scalability, rather than follow an IRL framework we consider a *direct imitation* framework where we attempt to directly learn a policy instead of the reward function and/or transition dynamics. Unlike inverse RL, this framework does not require an exact dynamic model nor an efficient planner. Rather, our approach requires only a simulator of the environment dynamics that is able to generate trajectories given a policy. Such a simulator is often available, even when a compact description of the transition dynamics and/or a planner are not.

Recent active learning work in the direct imitation framework includes confidence based autonomy (Chernova and Veloso, 2009), and the related dogged learning framework (Grollman and Jenkins, 2007), where a policy is learned as it is executed. When the learner is uncertain about what to do at a state, the policy is paused and the expert is queried about what action to take, resulting in a policy update. One difficulty in applying this approach is setting the uncertainty threshold for querying the expert. While an automated threshold selection approach is suggested (Chernova and Veloso, 2009), our experiments show that it is not always effective. Like our work, this approach requires a dynamics simulator.

Recently, Ross and Bagnell (2010); Ross et al. (2011) proposed algorithms for imitation learning that are able to actively query the expert at particular states during policy execution. They show that under certain assumptions these algorithms have better theoretical performance guarantees than pure passive imitation learning. Unfortunately, these algorithms query the expert quite aggressively making them impractical for human experts or computationally expensive with automated experts. In contrast, our work focuses on active querying for the purpose of minimizing the expert's labeling effort. Like our work, they also require a dynamics simulator to help select queries.

## 3 Problem Setup and Background

We consider imitation learning in the framework of Markov decision processes (MDPs). An MDP is a tuple $\langle S, A, T, R, I \rangle$, where $S$ is the set of states, $A$ is the finite set of actions, $T(s, a, s')$ is the transition function denoting the probability of transitioning to state $s'$ upon taking action $a$ in state $s$, $R(s) \in [0, 1]$ is the reward function giving the immediate reward in state $s$, and $I$ is the initial state distribution. A stationary policy $\pi : S \mapsto A$ is a deterministic mapping from states to actions such that $\pi(s)$ indicates the action to take in state $s$ when executing $\pi$. A non-stationary policy is a tuple $\pi = (\pi_1, \ldots, \pi_T)$ of $T$ stationary policies such that $\pi(s, t) = \pi_t(s)$ indicates the action to take in state $s$ and at time $t$ when executing $\pi$, where $T$ is the time horizon. The expert's policy, which we assume is deterministic, is denoted as $\pi^*$.

The $T$-horizon value of a policy $V(\pi)$ is the expected total reward of trajectories that start in $s_1 \sim I$ at time $t = 1$ and then execute $\pi$ for $T$ steps. We use $d_\pi^t$ to denote the state distribution induced at time step $t$ by starting in $s_1 \sim I$ and then executing $\pi$. Note that $d_\pi^1 = I$ for all policies. We use $d_\pi = \frac{1}{T} \sum_{t=1}^{T} d_\pi^t$ to denote the state distribution induced by policy $\pi$ over $T$ time steps. To sample an $(s, a)$ pair from $d_\pi^t$, we start in $s_1 \sim I$, execute $\pi$ to generate a trajectory $\mathcal{T} = (s_1, a_1, \ldots, s_T, a_T, s_{T+1})$ and set $(s, a) = (s_t, a_t)$. Similarly, to sample from $d_\pi$, we first sample a random time step $t \in \{1, \ldots, T\}$, and then sample an $(s, a)$ pair from $d_\pi^t$. Note that in order to sample from $d_{\pi^*}$ (or $d_{\pi^*}^t$), we need to execute $\pi^*$. Throughout the paper, we assume that the only way $\pi^*$ can be executed is by querying the expert for an action in the current state and executing the given action, which puts significant burden on the expert.

The regret of a policy $\pi$ with respect to an expert policy $\pi^*$ is equal to $V(\pi^*) - V(\pi)$. In imitation learning, the goal is to learn a policy $\pi$ from a hypothesis class $H$ (e.g. linear action classifiers), that has a small regret. In the passive setting of imitation learning, the learner is provided with a training set of full execution trajectories of $\pi^*$ and the state-action pairs (or a sample of them) are passed to an i.i.d. supervised learning algorithm. To help avoid the cost of generating full trajectories, active imitation learning allows the learner to pose *action queries*. In an action query, a state $s$ is presented to the expert and the expert returns the desired action $\pi^*(s)$.

In addition to having access to the expert for answering queries, we assume that the learner has access to a simulator of the MDP. The input to the simulator is a policy $\pi$ and a horizon $T$. The simulator output is a state trajectory that results from executing $\pi$ for $T$ steps starting in the initial state. The learner is allowed to interact with this simulator as part of its query selection process. The simulator is not assumed to provide a reward signal, which means that the learner cannot find $\pi$ by pure reinforcement learning.

Since our analysis in the next two sections is based on reducing to i.i.d. active learning and comparing to i.i.d. passive learning, we briefly review the *Probably Approximately Correct (PAC)* (Valiant, 1984) learning formulation for the i.i.d. setting. Here we consider the realizable PAC setting, which will be the focus of our initial analysis. Section 4.3, extends to the non-realizable, or agnostic setting. In passive i.i.d. supervised learning, $N$ i.i.d. data samples are drawn from an unknown distribution $D_\mathcal{X}$ over an input space $\mathcal{X}$ and are labeled according to an unknown target classifier $f : \mathcal{X} \mapsto \mathcal{Y}$, where $\mathcal{Y}$ denotes the label space. In the realizable PAC setting it is assumed that $f$ is an element of a known class of classifiers $H$ and given a set of $N$ examples a learner then outputs a hypothesis $h \in H$. Let $e_f(h, D_\mathcal{X}) = \mathbb{E}_{x \sim D_\mathcal{X}}[h(x) \neq f(x)]$ denote the generalization error of the returned classifier $h$. Standard PAC learning theory provides a bound on the number of labeled examples that are sufficient to guarantee that for any distribution $D_\mathcal{X}$, with probability at least $1 - \delta$, the returned classifier $h$ will satisfy $e_f(h, D_\mathcal{X}) \leq \epsilon$. We will denote this bound by $N_p(\epsilon, \delta)$, which corresponds to the label/query complexity of i.i.d. passive supervised learning for a class $H$. We will also denote a passive learner that achieves this label complexity as $L_p(\epsilon, \delta)$.

In i.i.d. active learning, the learner is given access to two resources rather than just a set of training data: 1) A "cheap" resource (Sample) that can draw an unlabeled sample from $D_\mathcal{X}$ and provide it to the learner when requested, 2) An "expensive" resource (Label) that can label a given unlabeled sample according to target concept $f$ when requested. Given access to these two resources, an active learning algorithm is required to learn a hypothesis $h \in H$ while posing as few queries to Label as possible. It can, however, pose a much larger number of queries to Sample (though still polynomial) as it is cheap. We use $N_a(\epsilon, \delta)$ to denote the label complexity (i.e. number of calls to Label) that is sufficient for an active learner to return an $h$ that for any $D_\mathcal{X}$ with probability at least $1 - \delta$ satisfies $e_f(h, D_\mathcal{X}) \leq \epsilon$. Similarly, we will denote an active learner that achieves this label complexity as $L_a(\epsilon, \delta, D)$, where the final argument $D$ indicates that the Sample function used by $L_a$ samples from distribution $D$. It has been shown that often $N_a$ can be exponentially smaller than $N_p$ for hypothesis classes with bounded complexity (e.g. VC-dimension). In particular, in the realizable case, ignoring $\delta$, $N_p = O(\frac{1}{\epsilon})$ whereas $N_a = O(\log(\frac{1}{\epsilon}))$ giving exponential improvement in label complexity over passive learning.

## 4 Reductions for Active Imitation Learning

We now consider reductions from active imitation learning to active i.i.d. learning for the cases of deterministic non-stationary and stationary policies.

### 4.1 Non-Stationary Policies

Syed and Schapire (2010) analyze the traditional reduction from passive imitation learning to passive i.i.d. learning for non-stationary policies. The algorithm uses queries to sample $N$ expert trajectories, noting that the state-action pairs at time $t$ can be viewed as i.i.d. draws from distribution $d_{\pi^*}^t$. The algorithm, then returns the non-stationary policy $\hat{\pi} = (\hat{\pi}_1, \ldots, \hat{\pi}_T)$, where $\hat{\pi}_t$ is the policy returned by running the learner $L_p$ on examples from time $t$. Let $\epsilon_t = e_{\pi_t^*}(\hat{\pi}_t, d_{\pi^*}^t)$ be the generalization error at time

$t$. Lemma 3[1] in (Syed and Schapire, 2010) shows that if for each time step $\epsilon_t \leq \epsilon$, then $V(\hat{\pi}) \geq V(\pi^*) - \epsilon T^2$. Hence, if we are interested in learning a $\hat{\pi}$ whose overall regret is no more than $\epsilon$, then we need to provide at least $N_p(\frac{\epsilon}{T^2}, \frac{\delta}{T})$ examples to $L_p$ to ensure that $\epsilon_t \leq \frac{\epsilon}{T^2}$ holds at all time steps with probability at least $1 - \delta$. Therefore, the passive label complexity of this algorithm is $T \cdot N_p(\frac{\epsilon}{T^2}, \frac{\delta}{T})$.

Our goal now is to provide a reduction from active imitation learning to i.i.d. active learning that can achieve an improved label complexity. This is not as simple as replacing the calls to $L_p$ in the above approach with calls to an active learner $L_a$. This is because the active learner at time step $t$ would require the ability to sample from the unlabeled distribution $d^t_{\pi^*}$, which requires executing the expert policy for $t$ steps, which in turn requires $t$ label queries to the expert that would count against the label complexity.

It turns out that for a slightly more sophisticated reduction to i.i.d. passive learning introduced by Ross and Bagnell (2010), it is possible to simply replace $L_p$ with $L_a$ and maintain the potential benefit of active learning. Ross and Bagnell (2010) introduced the forward training algorithm for non-stationary policies, which trains a non-stationary policy in a series of $T$ iterations. In particular, iteration $t$ trains policy $\hat{\pi}_t$ by calling a passive learner $L_p$ on a labeled data set drawn from the state distribution induced at time $t$ by the non-stationary policy $\hat{\pi}^{t-1} = (\hat{\pi}_1, \ldots, \hat{\pi}_{t-1})$, where $\hat{\pi}_1$ is learned on states drawn from the initial distribution $I$. The motivation for this approach is to train the policy at time step $t$ based on the same state-distribution that it will encounter when being run after learning. By doing this, Ross and Bagnell (2010) show that the algorithm has a worst case regret of $T^2\epsilon$ and under certain assumptions can achieve a regret as low as $O(T\epsilon)$.

Importantly, the state-distribution used to train $\hat{\pi}_t$ given by $d^t_{\hat{\pi}^{t-1}}$ is easy for the learner to sample from without making queries to the expert. In particular, to generate a sample the learner can simply simulate $\hat{\pi}^{t-1}$, which is available from previous iterations, from a random initial state and return the state at time $t$. Thus, we can simply replace the call to $L_p$ at iteration $t$ with a call to $L_a$ with unlabeled state distribution $d^t_{\hat{\pi}^{t-1}}$ as input. More formally, the *active forward training algorithm* is given by the following iteration: $\hat{\pi}_t = L_a(\epsilon, \frac{\delta}{T}, D^t)$, where $D^1 = I$ and $D^t = d^t_{\hat{\pi}^{t-1}}$.

Theorem 3.1 in (Ross and Bagnell, 2010) gives the worst case bound on the regret of the forward training algorithm which assumes the generalization error at each iteration is bounded by $\epsilon$. Since we also maintain that assumption when replacing $L_p$ with $L_a$ (the active variant) we immediately inherit that bound.

**Proposition 1.** *Given a PAC i.i.d. active learning algorithm $L_a$, if active forward training is run by giving $L_a$ parameters $\epsilon$ and $\frac{\delta}{T}$ at each step, then with probability at least $1 - \delta$ it will return a non-stationary policy $\hat{\pi}^T$ such that $V(\hat{\pi}^T) \geq V(\pi^*) - \epsilon T^2$.*

Note that $L_a$ is run with $\frac{\delta}{T}$ as the reliability parameter to ensure that all $T$ iterations succeed with the desired probability. Proposition 1 shows that the overall label complexity of active forward training in order to achieve a regret less than $\epsilon$ with probability at least $1 - \delta$ is $T \cdot N_a(\frac{\epsilon}{T^2}, \frac{\delta}{T})$. Comparing to the corresponding label complexity of passive imitation learning $T \cdot N_p(\frac{\epsilon}{T^2}, \frac{\delta}{T})$ we see that an improved label complexity of active learning in the i.i.d. case translates to improved label complexity of active imitation learning. In particular, in the realizable learning case, we know that $N_a$ can be exponentially smaller than $N_p$ in terms of its dependence on $\frac{1}{\epsilon}$.

### 4.2 Stationary Policies

A drawback of active forward training is that it is impractical for large $T$ and the resulting policy cannot be run indefinitely. We now consider the case of learning stationary policies, first reviewing the existing results for passive imitation learning.

In the traditional approach, a stationary policy $\hat{\pi}$ is trained on the expert state distribution $d_{\pi^*}$ using a passive learning algorithm $L_p$ returning a stationary policy $\hat{\pi}$. Theorem 2.1 in (Ross and Bagnell, 2010) states that if the generalization error of $\hat{\pi}$ with respect to the i.i.d. distribution $d_{\pi^*}$ is bounded by $\epsilon$ then $V(\hat{\pi}) \geq V(\pi^*) - \epsilon T^2$. Since generating i.i.d. samples from $d_{\pi^*}$ can require up to $T$ queries (see Section 3) the passive label complexity of this approach for guaranteeing a regret less than $\epsilon$ with probability at least $1 - \delta$ is $T \cdot N_p(\frac{\epsilon}{T^2}, \delta)$.

The above approach cannot be converted into an active imitation learner by simply replacing the call to $L_p$ with $L_a$, since again we cannot sample from the unlabeled distribution $d_{\pi^*}$ without querying the expert. To address this issue, we introduce a new algorithm called *RAIL (Reduction-based Active Imitation Learning)* which makes a sequence of $T$ calls to an i.i.d. active learner, noting that it is likely to find a useful stationary policy well before all $T$ calls are issued. RAIL is an idealized algorithm intended for analysis, which achieves the theoretical goals but has a number of inefficiencies from a practical perspective. Later in Section 5 we describe the practical instantiation that is used in our experiments.

---
[1] The main result of (Syed and Schapire, 2010) holds for stochastic expert policies and requires a more complicated analysis that results in a looser bound. Lemma 3 is strong enough for deterministic expert policies, which is the assumption made in our work.

RAIL is similar in spirit to active forward training, though its analysis is quite different and more involved. Like forward-training RAIL iterates for $T$ iterations, but on each iteration, RAIL learns a new stationary policy $\hat{\pi}^t$ that can be applied across all time steps. Iteration $t+1$ of RAIL learns a new policy $\hat{\pi}^{t+1}$ that achieves a low error rate at predicting the expert's actions with respect to the state distribution of the previous policy $d_{\hat{\pi}^t}$. More formally, given an i.i.d. active learner $L_a$, the RAIL algorithm is defined by the following iteration:

- **(Initialize)** $\hat{\pi}^0$ is an arbitrary policy, possibly based on prior knowledge or existing data,
- **(Iterate** $t = 1 \cdots, T$**)** $\hat{\pi}^t = L_a(\epsilon, \frac{\delta}{T}, d_{\hat{\pi}^{t-1}})$.

Thus, similar to active forward training, RAIL makes a sequence of $T$ calls to an active learner. Unlike forward training, however, the unlabeled data distributions used at each iteration contains states from all time points within the horizon, rather than being restricted to a states arising at a particular time point. Because of this difference, the active learner is able to ask queries across a range of time point and we might expect policies learned in earlier iterations to achieve non-trivial performance throughout the entire horizon. In contrast, at iteration $t$ the policy produced by forward training is only well defined up to time $t$.

The complication faced by RAIL, however, compared to forward training, is that the distribution used to train $\hat{\pi}^{t+1}$ differs from the state distribution of the expert policy $d_{\pi^*}$. This is particularly true in early iterations of RAIL since $\hat{\pi}^0$ is initialized arbitrarily. We now show that as the iterations proceed, we can bound the similarity between the state distributions of the learned policy and the expert, which allows us to bound the regret of the learned policy. We first state the main result which we prove below.

**Theorem 1.** *Given a PAC i.i.d. active learning algorithm $L_a$, if RAIL is run with parameters $\epsilon$ and $\frac{\delta}{T}$ passed to $L_a$ at each iteration, then with probability at least $1 - \delta$ it will return a stationary policy $\hat{\pi}^T$ such that $V(\hat{\pi}^T) \geq V(\pi^*) - \epsilon T^3$.*

From this we see that the impact of moving from non-stationary to stationary policies in the worst case is a factor of $T$ in the regret bound. Similarly the bound is a factor of $T$ worse than the comparable result above for passive imitation learning, which suffered a worst-case regret of $\epsilon T^2$. From this we see that the total label complexity for RAIL required to guarantee a regret of $\epsilon$ with probability $1 - \delta$ is $T \cdot N_a(\frac{\epsilon}{T^3}, \frac{\delta}{T})$ compared to the above label complexity of passive learning $T \cdot N_p(\frac{\epsilon}{T^2}, \delta)$. Thus, for a given policy class, if the label complexity of i.i.d. active learning is substantially less than the label complexity of passive learning, then our reduction can leverage those savings. For example, in the realizable learning case, ignore the dependence on $\delta$ (which is only logarithmic), we get an active label complexity of $T \cdot O(\log \frac{T^3}{\epsilon})$ versus the corresponding passive complexity of $T \cdot O(\frac{T^2}{\epsilon})$.

For the proof we introduce the quantity $P_\pi^t(M)$, which is the probability that a policy $\pi$ is consistent with a length $t$ trajectory generated by the expert policy $\pi^*$ in MDP $M$. It will also be useful to index the state distribution of $\pi$ by the MDP $M$, denoted by $d_\pi(M)$. The main idea is to show that at iteration $t$, $P_{\hat{\pi}^t}^t(M)$ is not too small, meaning that the policy at iteration $t$ mostly agrees with the expert for the first $t$ actions. We first state two lemmas, which are useful for the final proof. First, we bound the regret of a policy in terms of $P_\pi^T(M)$.

**Lemma 1.** *For any policy $\pi$, if $P_\pi^T(M) \geq 1 - \epsilon$, then $V(\pi) \geq V(\pi^*) - \epsilon T$.*

*Proof.* Let $\Gamma^*$ and $\Gamma$ be all state-action sequences of length $T$ that are consistent with $\pi^*$ and $\pi$ respectively. If $R(\mathcal{T})$ is the total reward for a sequence $\mathcal{T}$ then we get the following:

$$\begin{aligned}
V(\pi) &= \sum_{\mathcal{T} \in \Gamma} \Pr(\mathcal{T} \mid M, \pi) R(\mathcal{T}) \\
&\geq \sum_{\mathcal{T} \in \Gamma \cap \Gamma^*} \Pr(\mathcal{T} \mid M, \pi) R(\mathcal{T}) \\
&= \sum_{\mathcal{T} \in \Gamma^*} \Pr(\mathcal{T} \mid M, \pi^*) R(\mathcal{T}) - \sum_{\mathcal{T} \in \Gamma^* - \Gamma} \Pr(\mathcal{T} \mid M, \pi^*) R(\mathcal{T}) \\
&= V(\pi^*) - \sum_{\mathcal{T} \in \Gamma^* - \Gamma} \Pr(\mathcal{T} \mid M, \pi^*) R(\mathcal{T}) \\
&\geq V(\pi^*) - T \cdot \sum_{\mathcal{T} \in \Gamma^* - \Gamma} \Pr(\mathcal{T} \mid M, \pi^*) \\
&\geq V(\pi^*) - \epsilon T
\end{aligned}$$

The last two inequalities follow since the reward for a sequence must be no more than $T$ and our assumption about $P_\pi^T(M)$. □

Next, we show how the value of $P_\pi^t(M)$ changes across one iteration of learning.

**Lemma 2.** *For any policies $\pi$ and $\hat{\pi}$ and $1 \leq t < T$, if $e_{\pi^*}(\hat{\pi}, d_\pi(M)) \leq \epsilon$, then $P_{\hat{\pi}}^{t+1}(M) \geq P_\pi^t(M) - T\epsilon$.*

*Proof.* We define $\hat{\Gamma}$ to be all sequences of state-action pairs of length $t+1$ that are consistent with $\hat{\pi}$. Also define $\Gamma$ to be all length $t+1$ state-action sequences that are consistent with $\pi$ on the first $t$ state-action pairs (so need not be consistent on the final pair). We also define $\hat{M}$ to be an MDP that is identical to $M$, except that the transition distribution of any state-action pair $(s, a)$ is equal to the transition distribution of action $\pi(s)$ in state $s$. That is, all actions taken in a state $s$ behave like the action selected by $\pi$ in $s$.

We start by arguing that if $e_{\pi^*}(\hat{\pi}, d_\pi(M)) \leq \epsilon$ then $P_{\hat{\pi}}^{t+1}(\hat{M}) \geq 1 - T\epsilon$, which relates our error assumption to the MDP $\hat{M}$. To see this note that for MDP $\hat{M}$, all policies including $\pi^*$, have state distribution given by $d_\pi$. Thus by the union bound $1 - P_{\hat{\pi}}^{t+1}(\hat{M}) \leq \sum_{i=1}^{t+1} \epsilon_i$, where $\epsilon_i$ is the error of $\hat{\pi}$ at predicting $\pi^*$ on distribution $d_\pi^i$. This sum is bounded by $T\epsilon$ since $e_{\pi^*}(\hat{\pi}, d_\pi(M)) = \frac{1}{T}\sum_{i=1}^{T}\epsilon_i$. Using this fact we can now derive the following:

$$\begin{aligned}
P_{\hat{\pi}}^{t+1}(M) &= \sum_{\mathcal{T} \in \hat{\Gamma}} \Pr(\mathcal{T} \mid M, \pi^*) \\
&\geq \sum_{\mathcal{T} \in \Gamma \cap \hat{\Gamma}} \Pr(\mathcal{T} \mid M, \pi^*) \\
&= \sum_{\mathcal{T} \in \Gamma} \Pr(\mathcal{T} \mid M, \pi^*) - \sum_{\mathcal{T} \in \Gamma - \hat{\Gamma}} \Pr(\mathcal{T} \mid M, \pi^*) \\
&= P_\pi^t(M) - \sum_{\mathcal{T} \in \Gamma - \hat{\Gamma}} \Pr(\mathcal{T} \mid M, \pi^*) \\
&= P_\pi^t(M) - \sum_{\mathcal{T} \in \Gamma - \hat{\Gamma}} \Pr(\mathcal{T} \mid \hat{M}, \pi^*) \\
&\geq P_\pi^t(M) - \sum_{\mathcal{T} \notin \hat{\Gamma}} \Pr(\mathcal{T} \mid \hat{M}, \pi^*) \\
&\geq P_\pi^t(M) - T\epsilon
\end{aligned}$$

The equality of the fourth line follows since $\Gamma$ contains all sequences whose first $t$ actions are consistent with $\pi$ with all possible combinations of the remaining action and state transition. Thus, summing over all such sequences yields the probability that $\pi^*$ agrees with the first $t$ steps. The equality of the fifth line follows because $\Pr(\mathcal{T} \mid M, \pi^*) = \Pr(\mathcal{T} \mid \hat{M}, \pi^*)$ for any $\mathcal{T}$ that is in $\Gamma$ and for which $\pi^*$ is consistent (has non-zero probability under $\pi^*$). The final line follows from the above observation that $P_{\hat{\pi}}^{t+1}(\hat{M}) \geq 1 - T\epsilon$. □

We can now complete the proof of the main theorem.

*Proof of Theorem 1.* Using failure parameter $\frac{\delta}{T}$ ensures that with at least probability $1 - \delta$ that for all $1 \leq t < T$ we will have $e_{\pi^*}(\hat{\pi}^{t+1}, d_{\hat{\pi}^t}(M)) \leq \epsilon$. As a base case, we have $P_{\hat{\pi}^1}^1 \geq 1 - T\epsilon$, since the the error rate of $\hat{\pi}^1$ relative to the initial state distribution at time step $t = 1$ is at most $T\epsilon$. Combining these facts with Lemma 2 we get that $P_{\hat{\pi}^T}^T \geq 1 - \epsilon T^2$. Combining this with Lemma 1 completes the proof. □

### 4.3 Agnostic Case

Above we considered the realizable setting, where the expert's policy was assumed to be in a known hypothesis class $H$. In the agnostic case, we do not make such an assumption. The learner still outputs a hypothesis from a class $H$, but the unknown policy is not necessarily in $H$. The agnostic i.i.d. PAC learning setting is defined similarly to the realizable setting, except that rather than achieving a specified error bound of $\epsilon$ with high probability, a learner must guarantee an error bound of $\inf_{\pi \in H} e_f(\pi, D_\mathcal{X}) + \epsilon$ with high probability (where $f$ is the target), where $D_\mathcal{X}$ is the unknown data distribution. That is, the learner is able to achieve close to the best possible accuracy given class $H$. In the agnostic case, it has been shown that exponential improvement in label complexity with respect to $\frac{1}{\epsilon}$ is achievable when $\inf_{\pi \in H} e_f(\pi, D_\mathcal{X})$ is relatively small compared to $\epsilon$ (Dasgupta, 2011). Further, there are many empirical results for practical active learning algorithms that demonstrate improved label complexity compared to passive learning.

It is straightforward to extend our above results for non-stationary and stationary policies to the agnostic case by using agnostic PAC learners for $L_p$ and $L_a$. Space precludes full details and here we outline the extension for RAIL. Note that the RAIL algorithm will call $L_a$ using a sequence of unlabeled data distributions, where each distribution is of the form $d_\pi$ for some $\pi \in H$ and each of which may yield a different minimum error given $H$. For this purpose, we define $\epsilon^* = \sup_{\pi \in H} \inf_{\hat{\pi} \in H} e_{\pi^*}(\hat{\pi}, d_\pi)$ to be the minimum generalization error achievable in the worst case considering all possible state distributions $d_\pi$ that RAIL might possibly encounter. With minimal changes to the proof of Theorem 1, we can get an identical result, except that the regret is $(\epsilon^* + \epsilon)T^3$ rather than just $\epsilon T^3$. A similar change in regret holds for passive imitation learning. This shows that in the agnostic setting we can get significant improvements in label complexity via active imitation learning when there are significant savings in the i.i.d. case.

## 5 Practical Instantiation of RAIL

Despite the theoretical guarantees, a potential drawback of RAIL is that the unlabeled state distributions used at early iterations may be quite different from $d_{\pi^*}$. In particular, at iteration $t$, the distributions at times $t' > t$ have no guarantees with respect to $d_{\pi^*}^{t'}$. Thus, early iterations may focus substantial query effort on parts of the state-space that are not relevant to $\pi^*$. Another issue is that the idealized version does not share data across iterations, which is potentially wasteful. We now describe our practical instantiation of RAIL, called RAIL-DW (for density weighted), which addresses these issues in several ways.

First, we use an incremental version of RAIL that asks only one query per iteration and accumulates data across iterations. This allows rapid updating of state distributions and prevents RAIL from wasting its query budget on earlier inaccurate distributions but rather focus on later more accurate distributions.

Second, recall that at iteration $t+1$, RAIL learns using an unlabeled data distribution $d_{\hat{\pi}^t}$, where $\hat{\pi}^t$ is the policy learned at iteration $t$. In order to help improve the accuracy of this unlabeled distribution (with respect to $d_{\pi^*}$), instead of using a point estimate for $\hat{\pi}^t$, we treat $\hat{\pi}^t$ as a Bayesian classifier for purposes of defining $d_{\hat{\pi}^t}$. In particular, at iteration $t$ let $D_t$ be the set of state-action pairs collected from previous iterations. We use this to define a posterior $P(\hat{\pi}|\mathcal{D})$ over policies in our policy class $H$. This, in turn, defines a posterior unlabeled state distribution $d_{D_t} = \mathbb{E}_{\hat{\pi} \sim P(\hat{\pi}|\mathcal{D})}[d_{\hat{\pi}}(s)]$ which is used in place of $d_{\hat{\pi}^t}$ in RAIL. Note that we can sample from this distribution by first sampling a policy $\hat{\pi}$ and then sampling a state from $d_{\hat{\pi}}$, all of which can be done without interaction with the expert. We observe that in practice $d_{D_t}$ is a significantly more useful estimate of $d_{\pi^*}$ than the point estimate, since its more highly weighted states tend to carry significant weight according to $d_{\pi^*}$.

To summarize, iteration $t$ of RAIL-DW carries out the following steps, adding one state-action pair to the data set: ($D_1$ is the initial, possibly empty, data set)

1. Run active learner $L_a$ using unlabeled data distribution $d_{D_t}$ to select a single query state $s$.

2. Query the expert about $s$ to obtain action $\pi^*(s)$ and let $D_{t+1} = D_t \cup \{(s, \pi^*(s))\}$.

The iteration repeats until the query budget is exceeded. It remains to specify the specific i.i.d. active learning algorithm that we use and how we sample from $d_{D_t}$.

Since it is important that the i.i.d. active learner be sensitive to the unlabeled data distribution, we employ a density-weighted learning algorithm (hence the name RAIL-DW). In particular, we use density-weighted query-by-committee (McCallum and Nigam, 1998) in our implementation. Given a sample of unlabeled data points, this approach uses bagging (Breiman, 1996) to generate a committee for query selection and also uses a density estimator to estimate the density of the unlabeled data points. The selected query is the state that maximizes the product of state density and committee disagreement. We use the entropy of the vote distribution (Dagan and Engelson, 1995) as a typical measure of committee disagreement. We use a committee of size 5 in our experiments and estimate density of unlabeled points via simple distance based binning.

Finally, for sampling we assume a class of linear parametric policies and assume a uniform prior over the parameters. We approximate sampling from $d_{D_t}$ via bagging: where we generate $K$ bootstrap samples of $D_t$ and a policy is learned from each using a supervised learner. This produces a set of $K$ policies and each is executed to generate $K$ trajectories. The states on those trajectories are given to the active learner as the unlabeled data set. We set $K = 5$ in our experiments.

## 6 Experiments

We empirically evaluate RAIL-DW on four domains: 1) Cart-pole, 2) Bicycle, 3) Wargus and 4) The structured prediction domain NETtalk. We compare RAIL-DW against the following baselines: 1) *Passive*, which simulates the traditional approach by starting at the initial state and querying the expert about what to do at each visited state, 2) *unif-QBC*, which views all the states as i.i.d. according to the uniform distribution and applies the standard query-by-committee (QBC) (Seung et al., 1992) active learning approach. Intuitively, this approach will select the state with highest action uncertainty according to the current data set and ignore the state distribution, 3) *unif-RAND*, which selects states to query uniformly at random, and 4) *Confidence based autonomy (CBA)* (Chernova and Veloso, 2009), which, starting at the initial state, executes the current policy until the learner's confidence falls below an automatically determined threshold at which point it queries the expert for an action. CBA may decide to stop asking queries once the confidence exceeds the threshold in all states. We use the exact automated threshold adjustment strategy proposed in (Chernova and Veloso, 2009). For all of these methods, we employed the SimpleLogistic classifier in Weka (Hall et al., 2009) to learn policies.

**Cart-Pole.** Cart-pole is an RL benchmark where a cart must balance an attached vertical pole by applying left or right forces to the cart. An episode ends when either the pole falls or the cart goes out of bounds. There are two actions, left and right, and four state variables describing the position and velocity of the cart and the angle and angular velocity of the pole. We made slight modifications to the usual setting where we allow the pole to fall down and become horizontal and the cart to go out of bounds (we used default [-2.4, 2.4] as the bounds for the cart). We let each episode run for a fixed length of 5000 time steps. This opens up the possibility of generating several "undesirable" states where either the pole has fallen or the cart is out of bounds that are rarely or never generated by the expert's state distribution. The expert policy was a hand-coded policy that can balance the pole indefinitely. For each learner, we ran experiments from 30 random initial states close to the equilibrium start state, generating learning curves for each one and averaging the results. We report total reward with a reward function (unknown to the learner) that is +1 for each time step where the pole is balanced and the cart is in bounds and -1 otherwise.

Figure 1(a) shows the results for cart-pole. We observe that RAIL learns quickly and achieves optimal performance with only 30-35 queries. Passive, on the other hand, takes 100 queries to get close to the optimal per-formance. The reason for this difference is clear when one visualizes the states queried by RAIL versus Passive (not shown for space reasons). The queries posed by RAIL tend to be close to the decision boundary of the expert policy, while passive asks many uninformative queries that are not close to the boundary. However, a naive reduction to active learning can be dangerous, as demonstrated by the poor performance of unif-QBC. Further, RAIL performs much better than random query selection as demonstrated by the performance of unif-RAND. By ignoring the real data distribution altogether and incorrectly assuming it to be uniform, these naive methods end up asking many queries that are not relevant to learning the expert policy (e.g. states where the pole is in an unrecoverable fall or the cart is out of bounds). CBA, like RAIL, learns quickly but settles at a suboptimal performance. This is because it becomes confident and stops asking queries prematurely. This shows that CBA's automatic threshold adjustment mechanism did not work well in this domain. We did try several adjustments to the threshold adjustment strategy, but were unable to find one that was robust across our four domains and thus we report results for the original strategy.

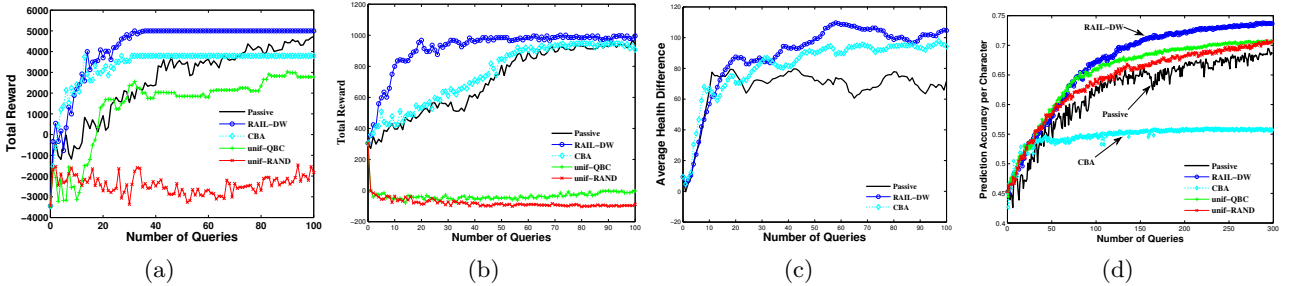

Figure 1: Active imitation learning results: (a) Cart-pole (b) Bicycle balancing (c) Wargus (d) NETtalk.

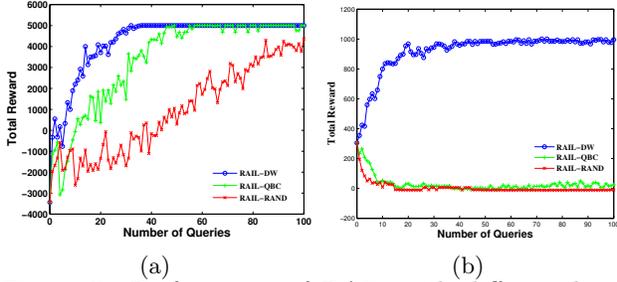

Figure 2: Performance of RAIL with different base active learners on (a) Cart-pole (b) Bicycle balancing.

We also conducted experiments to study the effects of using i.i.d. active learners that ignore the data distribution in RAIL. Figure 2(a) shows the performance of RAIL with three different base active learners, the density-weighted QBC, the standard QBC (without density weighting), and random selection of unlabeled data points. We see that RAIL-DW performs better than both RAIL-QBC and RAIL-RAND. This shows that it is critical for the i.i.d. active learner to exploit the state density information that is estimated by RAIL at each iteration.

**Bicycle Balancing.** Bicycle balancing is a variant of the Bicycle RL benchmark (Randløv and Alstrøm, 1998). The goal is to balance a bicycle moving at a constant speed for 1000 time steps. If the bicycle falls, it remains fallen for the rest of the episode. The state space is described using nine variables measuring various angles, angular velocities, and positions of the bicycle. There are five possible actions each specifying a particular handle-bar torque and rider displacement. The learner's policy is represented as a linear logistic regression classifier over features of state-action pairs. A feature vector is defined as follows: Given a state $s$, a set of 20 basis functions is computed. This set is repeated for each of the 5 actions giving a feature vector of length 100. The expert policy was hand coded and can balance the bicycle for up to 26K time steps.

We used a similar evaluation procedure as for Cart-pole giving +1 reward for each time step where the bicycle is kept balanced and -1 otherwise. Figure 1(b) compares each approach. The results are similar to those of Cart-pole with RAIL-DW being the top performer. Unif-RAND and Unif-QBC show notably poor performance in this domain. This is because bicycle balancing is a harder learning problem than cart-pole with many more uninformative states (an unrecoverable fall or fallen state). Similarly, 2(b) shows very poor performance for versions of RAIL that use distribution unaware active learners.

**Wargus.** We consider controlling a group of 5 friendly close-range military units against a group of 5 enemy units in the real-time strategy game Wargus, similar to the setup in (Judah et al., 2010). The objective is to win the battle while minimizing the loss in total health of friendly units. The set of actions available to each friendly unit is to attack any one of the remaining units present in the battle (including other friendly units, which is always a bad choice). In our setup, we allow the learner to control one of the units throughout the battle whereas the other friendly units are controlled by a fixed "reasonably good" policy. This situation would arise when training the group via coordinate ascent on the performance of individual units. The expert policy corresponds to the same policy used by the other units. Note that poor behavior from even a single unit generally results in a huge loss. Providing full demonstrations in real time in such tactical battles is very difficult for human players and quite time consuming if demonstrations are done in slow motion, which motivates state-based active learning.

We designed 21 battle maps differing in the initial unit positions, using 5 for training and 16 for testing. We average results across five learning trials. Passive learns along the expert's trajectory in each map on all 5 maps considered sequentially according to a random ordering. For RAIL, in each iteration a training map is selected randomly and a query is posed in the chosen map. For CBA, a map is selected randomly and CBA is allowed to play an episode in it, pausing and querying as and when needed. If the episode ends, another map is chosen randomly and CBA continues to learn in it. After each query, the learned policy is tested on the 16 test maps. We use the difference in the total health of friendly and enemy units at the end of the battle as the performance metric (which is positive for a win). We did not run experiments for unif-QBC and unif-RAND because it is difficult to define the space of feasible states over which to sample uniformly.

The results are shown in figure 1(c). We see that although Passive learns quickly for the first 10 queries, it fails to improve further. This shows that the states located in this initial prefix of the expert's trajectory are very useful, but thereafter Passive gets stuck on the uninformative part of the trajectory till its query budget is over. On the other hand, RAIL and CBA continue to improve beyond Passive, with RAIL being slightly better, which indicates that they are able to locate and query more informative states.

**NETTalk.** We evaluate RAIL on a structured prediction task of stress prediction in the NETtalk data set (Dietterich et al., 2008). Given a word, the goal is to assign one of the five stress labels to each letter of the word in the left to right order. It is straightforward to view structured prediction as imitation learning (see for example Ross and Bagnell, 2010) where at each time step (letter location), the learner has to execute the correct action (i.e. predict correct stress label) given the current state. The state consists of features describing the input (the current letter and its immediate neighbors) and previous $L$ predictions made by the learner (the prediction context). In our experiments, we use $L = 1, 2$ and show results only for $L = 2$ noting that $L = 1$ was qualitatively similar.

We use character accuracy as a measure of performance. Passive learns along the expert's trajectory on each training sequence considered in the order it appears in the training set. Therefore, Passive always learns on the correct context, i.e. previous $L$ characters correctly labeled. RAIL and Unif-QBC can select the best query across the entire training set. CBA, like Passive, considers training sequences in the order they appear in the training set and learns on each sequence by querying at each sequence position. In order to minimize the impact of the ordering of training examples on Passive and CBA, we ran five learning trials for each learner, each with a randomized ordering. We report final performance as the learning curves averaged across all five trials.

Figure 1(d) presents the NETtalk results. The results are qualitatively similar to Cart-Pole, except that Unif-QBC and Unif-RAND do quite well in this domain but not as well as RAIL-DW. Again we see that CBA performs poorly because in all five trials it prematurely stops asking queries, showing its sensitivity to its threshold adjustment mechanism.

**Overall Observations.** Overall, we can draw a number of conclusions from the experiments. First, RAIL-DW proved to be the most robust and effective active imitation learning approach in all of our domains. Second, the choice of the i.i.d. active learner used for RAIL is important. In particular, performance can be poor when the active learning algorithm does not take density information into account. Using density weighted query-by-committee was effective in all of our domains. Third, we found that CBA is quite sensitive to the threshold adjustment mechanism and we were unable to find an alternative mechanism that works across our domains. Fourth, we showed that a more naive application of i.i.d. active learning in the imitation setting is not effective.

## 7 Summary

We considered reductions from active imitation learning based on i.i.d. active learning, which allow for advances in the i.i.d. setting to translate to imitation learning. First, we analyzed the label complexity of reductions for both non-stationary and stationary policies, showing that the complexity of active imitation learning can be significantly less than passive learning. Second, we introduced RAIL-DW, a practical variant of the reduction for stationary policies, and showed empirically in four domains that it is highly effective compared to a number of alternatives.

**Acknowledgements**

The authors acknowledge support of the ONR ATL program N00014-11-1-0105.

# References

L. Breiman. Bagging predictors. *Machine learning*, 1996.

S. Chernova and M. Veloso. Interactive policy learning through confidence-based autonomy. *JAIR*, 2009.

J.A. Clouse. An introspection approach to querying a trainer. Technical report, University of Massachusetts, 1996.

R. Cohn, M. Maxim, E. Durfee, and S. Singh. Selecting operator queries using expected myopic gain. In *2010 IEEE/WIC/ACM International Conference on Web Intelligence and Intelligent Agent Technology*, 2010.

I. Dagan and S.P. Engelson. Committee-based sampling for training probabilistic classifiers. In *ICML*, 1995.

Sanjoy Dasgupta. Two faces of active learning. *Theor. Comput. Sci.*, 2011.

T. Dietterich, G. Hao, and A. Ashenfelter. Gradient tree boosting for training conditional random fields. *JMLR*, 2008.

F. Doshi, J. Pineau, and N. Roy. Reinforcement learning with limited reinforcement: Using Bayes risk for active learning in POMDPs. In *ICML*, 2008.

A. Gil, H. Stern, and Y. Edan. A Cognitive Robot Collaborative Reinforcement Learning Algorithm. *World Academy of Science, Engineering and Technology*, 2009.

D.H. Grollman and O.C. Jenkins. Dogged learning for robots. In *ICRA*, 2007.

M. Hall, E. Frank, G. Holmes, B. Pfahringer, P. Reutemann, and I.H. Witten. The WEKA data mining software: an update. *ACM SIGKDD Explorations Newsletter*, 2009.

Kshitij Judah, Saikat Roy, Alan Fern, and Thomas Dietterich. Reinforcement learning via practice and critique advice. In *AAAI*, 2010.

Manuel Lopes, Francisco S. Melo, and Luis Montesano. Active learning for reward estimation in inverse reinforcement learning. In *Principles of Data Mining and Knowledge Discovery*, 2009.

Andrew McCallum and Kamal Nigam. Employing em and pool-based active learning for text classification. In *Proceedings of the Fifteenth International Conference on Machine Learning*, 1998.

L. Mihalkova and R. Mooney. Using active relocation to aid reinforcement learning. In *FLAIRS*, 2006.

A.Y. Ng and S. Russell. Algorithms for inverse reinforcement learning. In *Proceedings of the Seventeenth International Conference on Machine Learning*, 2000.

J. Randløv and P. Alstrøm. Learning to drive a bicycle using reinforcement learning and shaping. In *Proceedings of the Fifteenth International Conference on Machine Learning*, 1998.

S. Ross and J.A. Bagnell. Efficient reductions for imitation learning. In *AISTATS*, 2010.

Stephane Ross, Geoffrey Gordon, and J. Andrew (Drew) Bagnell. A reduction of imitation learning and structured prediction to no-regret online learning. In *AISTATS*, 2011.

Burr Settles. Active learning literature survey. Technical report, University of Wisconsin–Madison, 2009.

H.S. Seung, M. Opper, and H. Sompolinsky. Query by committee. In *COLT*, 1992.

A.P. Shon, D. Verma, and R.P.N. Rao. Active imitation learning. In *AAAI*, 2007.

Umar Syed and Robert Schapire. A reduction from apprenticeship learning to classification. In *NIPS*, 2010.

L. G. Valiant. A theory of the learnable. *Commun. ACM*, 1984.